\documentclass[sigconf]{acmart}
\AtBeginDocument{%
  }

\copyrightyear{2026}
\acmYear{2026}
\setcopyright{cc}
\setcctype{by}
\acmConference[KDD '26]{Proceedings of the 32nd ACM SIGKDD Conference on Knowledge Discovery and Data Mining V.2}{August 09--13, 2026}{Jeju Island, Republic of Korea}
\acmBooktitle{Proceedings of the 32nd ACM SIGKDD Conference on Knowledge Discovery and Data Mining V.2 (KDD '26), August 09--13, 2026, Jeju Island, Republic of Korea}
\acmDOI{10.1145/3770855.3817964}
\acmISBN{979-8-4007-2259-2/2026/08}

\acmSubmissionID{2rtp2202}

\usepackage{style}
\usepackage[ruled,vlined]{algorithm2e}
\usepackage{multirow}
\usepackage{microtype}
\usepackage[htt]{hyphenat}
\usepackage{cuted}
\usepackage{float}
\usepackage{caption}

\begin{document}

% \title{Search-on-Graph: Reasoning on Knowledge Graphs as Informed Navigation by Large Language Model}
\title{Search-on-Graph: Iterative Informed Navigation for Large Language Model Reasoning on Knowledge Graphs}

\author{Jia Ao Sun}
\authornote{Equal contribution.}
\orcid{0000-0002-8340-155X}
\affiliation{%
  \institution{Université de Montréal}
  \city{Montréal}
  \state{Québec}
  \country{Canada}
}
\email{jia.ao.sun@umontreal.ca}

\author{Hao Yu}
\authornotemark[1]
\orcid{0009-0000-1770-2904}
\affiliation{%
  \institution{McGill University}
  \city{Montréal}
  \state{Québec}
  \country{Canada}
}
\email{hao.yu2@mail.mcgill.ca}

\author{Fabrizio Gotti}
\orcid{0000-0002-2602-2019}
\affiliation{%
  \institution{Digital Media, CBC}
  \city{Montréal}
  \state{Québec}
  \country{Canada}
}
\email{fabrizio.gotti@cbc.ca}

\author{Fengran Mo}
\orcid{0000-0002-0838-6994}
\affiliation{%
  \institution{Université de Montréal}
  \city{Montréal}
  \state{Québec}
  \country{Canada}
}
\email{fengran.mo@umontreal.ca}

\author{Yihong Wu}
\orcid{0009-0009-2680-4107}
\affiliation{%
  \institution{Université de Montréal}
  \city{Montréal}
  \state{Québec}
  \country{Canada}
}
\email{yihong.wu@umontreal.ca}

\author{Yuchen Hui}
\orcid{0000-0002-9659-3714}
\affiliation{%
  \institution{Université de Montréal}
  \city{Montréal}
  \state{Québec}
  \country{Canada}
}
\email{yuchen.hui@umontreal.ca}

\author{Zhan Su}
\orcid{0000-0001-5189-9165}
\affiliation{%
  \institution{Halmstad University College}
  \city{Halmstad}
  \state{}
  \country{Sweden}
}
\email{zhan.su@hh.se}

\author{Lingfeng Xiao}
\orcid{0009-0009-6101-360X}
\affiliation{%
  \institution{University of Waterloo}
  \city{Waterloo}
  \state{Ontario}
  \country{Canada}
}
\email{l7xiao@uwaterloo.ca}

\author{Jian-Yun Nie}
\authornote{Corresponding author.}
\orcid{0000-0003-1556-3335}
\affiliation{%
  \institution{Université de Montréal}
  \city{Montréal}
  \state{Québec}
  \country{Canada}
}
\email{nie@iro.umontreal.ca}

\renewcommand{\shortauthors}{Sun et al.}

\begin{abstract}
Large language models (LLMs) augmented with knowledge graphs (KGs) offer a promising approach for knowledge-intensive reasoning. Central to this approach is the selection of appropriate reasoning paths in the KG. Yet, existing methods face a common limitation: reasoning path selection is often performed by separate modules using criteria that are only weakly connected to the reasoning requirements. This often results in selecting incorrect relations or premature pruning of relevant paths. We propose Search-on-Graph (SoG), a method that strengthens the connection between path selection and reasoning by having the LLM itself select which relations to follow, informed by both the available KG structure and the complete reasoning history. SoG follows an \textit{observe-think-navigate} paradigm: at each step, the LLM observes the relational connections available at the current entity, reasons about which path best advances toward answering the question, and navigates accordingly. This context-aware navigation fully exploits the LLM's reasoning capabilities rather than relying on independent selection modules with surrogate criteria. Experiments on six knowledge graph question answering (KGQA) benchmarks demonstrate that SoG outperforms state-of-the-art methods while requiring no task-specific fine-tuning and generalizing across different KG schemas.
\end{abstract}





\maketitle

\section{Introduction}

Large language models (LLMs) have demonstrated remarkable capabilities across diverse natural language processing tasks through extensive pre-training on vast text corpora \citep{brown2020language, kojima2022large, wei2022chain, dubey2024llama}. However, these models face critical limitations when confronted with knowledge-intensive reasoning tasks. They hallucinate plausible-sounding but factually incorrect statements \citep{tonmoy2024comprehensive, huang2025survey}, operate with parametric knowledge that becomes rapidly outdated \citep{liska2022streamingqa, kasai2023realtime}, and lack the specialized domain expertise required for technical fields \citep{singhal2023large, kandpal2023large}. These limitations are particularly acute in multi-hop reasoning scenarios, where each reasoning step depends on accurate knowledge retrieval and where errors compound across the reasoning chain \citep{lightman2023let, ling2023deductive}. Such weaknesses significantly undermine the reliability and trustworthiness of LLMs in real-world applications that demand both factual accuracy and complex reasoning.

To address these challenges, augmenting LLMs with knowledge graphs (KGs) has emerged as a promising approach \citep{sun2024thinkongraph, chen2024plan, zhu2025beyond}. KGs encode factual relationships as typed edges between entities, providing structured knowledge that can be traced, verified, and updated independently of model parameters. However, developing generalizable methodologies for knowledge graph question answering (KGQA) remains difficult due to the scale, structural complexity, and schema heterogeneity of KGs such as Freebase %\citep{bollacker2008freebase} 
and Wikidata.
%\citep{vrandevcic2014wikidata}.

\begin{figure*}[t]
\begin{center}
  \includegraphics[width=\textwidth]{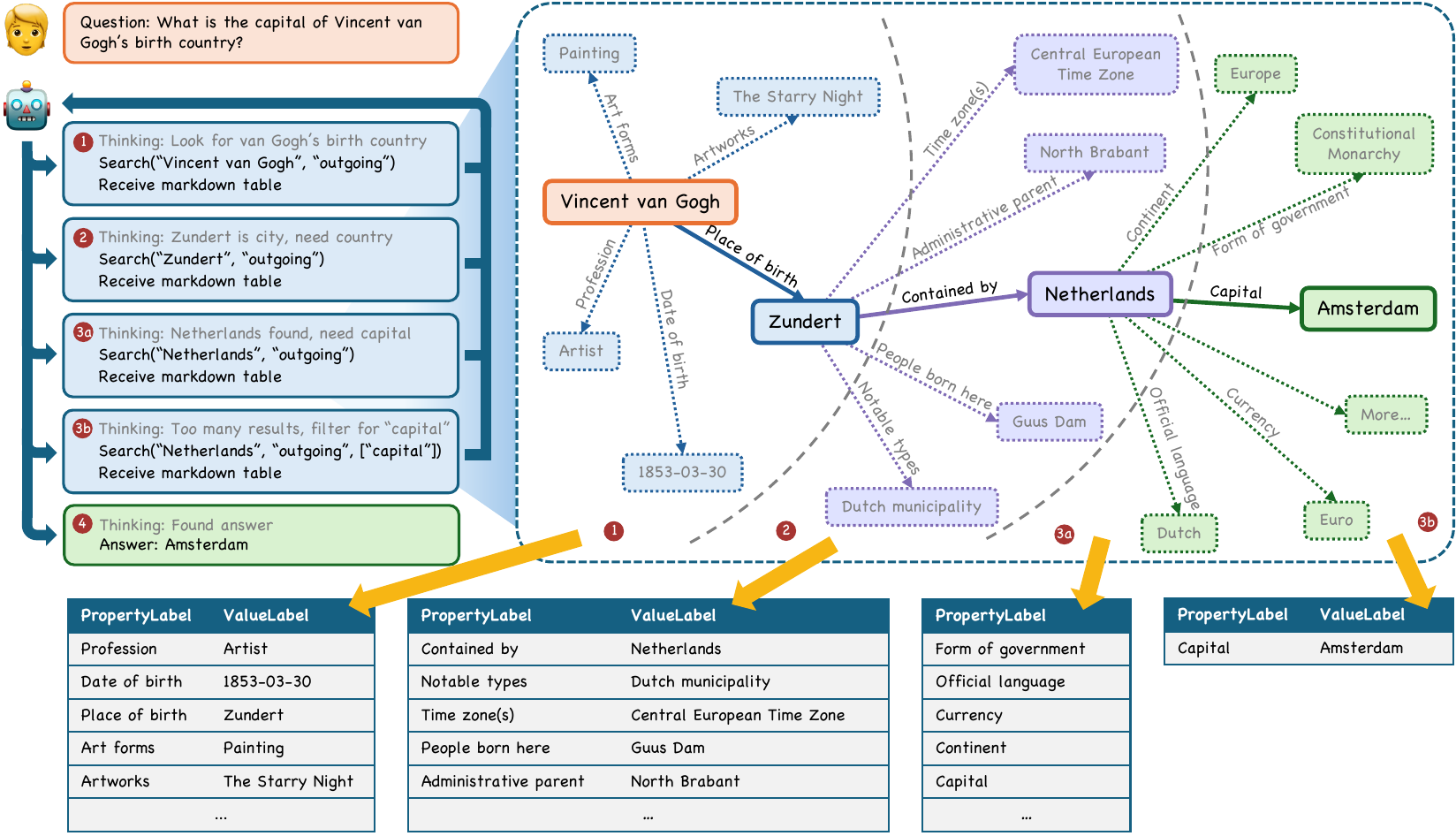}
  \label{fig:sog_workflow}
\end{center}
\caption{SoG workflow for the query ``\textit{What is the capital of Vincent van Gogh's birth country?}'' The LLM iteratively navigates the KG, with reasoning and \textsc{Search} calls shown on the left and KG navigation on the right. The path follows Van Gogh $\rightarrow$ Zundert (place of birth) $\rightarrow$ Netherlands (contained by) $\rightarrow$ Amsterdam (capital). Solid boxes indicate selected entities; dotted boxes show unselected retrieved entities. Tables display the markdown output returned by \textsc{Search}, revealing available 1-hop neighbours from each entity.}
\Description{A diagram showing the Search-on-Graph workflow. On the left side, an LLM generates reasoning steps and Search function calls. On the right side, a knowledge graph shows the navigation path from Vincent van Gogh to Zundert to Netherlands to Amsterdam, with solid boxes for selected entities and dotted boxes for unselected candidates. Markdown tables show the 1-hop neighbors retrieved at each step.}
\label{fig:sog-workflow}
\end{figure*}

While semantic parsing and subgraph retrieval methods have made progress on KGQA (see Section~\ref{sec:related-work}), recent agentic LLM approaches that iteratively explore KGs have emerged as a promising direction \citep{ma2025debate,jiang2025kg,wang2025reasoning}. However, existing methods have notable limitations. Some require multi-component architectures with separate modules for planning, retrieval, and reasoning, each relying on different criteria %where errors can propagate across components 
\citep{sui2024fidelis,luo2024graph,wang2025iquest,liu2025ontology}. Others depend on comprehensive upfront planning \citep{chen2024plan,li2024framework,cheng2024call}, which fails when presumed relations are absent from the actual KG. Methods like Think-on-Graph (ToG) \citep{sun2024thinkongraph} and EffiQA \citep{dong-etal-2025-effiqa} perform context-free pruning: ToG scores candidate relations based on semantic similarity to the original question, while EffiQA classifies entity types against pre-planned expectations—both ignoring the reasoning history that led to the current entity. This causes semantic drift, where exploration diverges from the logical chain required to answer the question, risking early elimination of logically correct but semantically dissimilar relations.

In response to these challenges, we propose Search-on-Graph (SoG), a fundamentally simpler methodology where a single LLM orchestrates iterative KG traversal through one carefully engineered \textsc{Search} function executing 1-hop exploration. SoG follows an \textit{observe-think-navigate} approach: rather than speculative path planning or semantic similarity heuristics, the LLM first systematically observes actual available relational connections at each entity, and then formulates informed navigational decisions grounded in question-specific reasoning. In particular, SoG performs context-aware navigation: each relation selection considers the full reasoning history (e.g., ``\textit{I am at Entity A, I came here from Entity B because I was looking for X, so now I should look for Y}''), ensuring logical coherence without parallel exploration.

For the query ``\textit{What is the capital of Vincent van Gogh's birth country?}'' illustrated in Figure~\ref{fig:sog-workflow}, the LLM executes iterative \textsc{Search} calls after observing relational options at each hop. From \textit{Van Gogh}, it identifies ``\textit{Place of birth}'' as relevant relation and navigates to Zundert, subsequently discovers ``\textit{Contained by}'' to reach \textit{Netherlands}, and finally selects ``\textit{Capital}'' to arrive at \textit{Amsterdam}. This methodology adapts to heterogeneous schemas---if \textit{Van Gogh} is connected directly to \textit{Netherlands} through a ``\textit{country of birth}'' relation in an alternative KG, the LLM would adopt the shorter path. %Our architectural simplicity stems from three deliberate design decisions: (1) an exploration function with compact result formatting that conserves context length, (2) a dynamic filtering mechanism that returns unique relation types when encountering large neighbourhoods, and (3) systematically engineered prompts that demonstrate effective reasoning processes. These seemingly simple design choices prove crucial across diverse KG schemas and question types.

% The key assumption behind this approach is that an LLM is more able to select the relevant relation to follow in the next step than a separate selection module. In particular, when making a selection, the LLM is more able to consider the reasoning needs (the reasoning logic), rather than relying on a semantic similarity measure. 
% As a matter of fact, using the representative ?? approach, the method would identify that ?? as a relevant node.
The key assumption behind this approach is that an LLM is better equipped to select the relevant relations at each step than a separate selection module. In particular, the LLM can base its selection on the actual reasoning requirements rather than relying on surface-level semantic similarity.

Empirically, SoG delivers strong and consistent gains across six KGQA benchmarks spanning Freebase and Wikidata. Our method achieves state-of-the-art performance on five of six datasets. The improvements are particularly notable on Wikidata-based benchmarks, where we achieve gains of over 11.3\% compared to previous best methods. %On Freebase datasets, SoG consistently outperforms existing approaches with meaningful improvements across different reasoning complexities.

Our main contributions are as follows:
\begin{itemize}
% \begin{sloppypar}
\item We propose Search-on-Graph (SoG), a simple yet effective KGQA framework that fully exploits the reasoning capability of LLMs and %where a single LLM navigates KGs through 
a 1-hop \textsc{Search} function is designed to inform the LLMs about the existing relations in KG. Unlike beam search methods that maintain parallel exploration paths with context-free pruning, SoG performs context-aware navigation based on the full reasoning history, avoiding semantic drift.
% \end{sloppypar}

\item We conduct extensive experiments across six KGQA benchmarks spanning Freebase and Wikidata, demonstrating that SoG achieves SOTA results without task-specific fine-tuning.

% \item Our study demonstrates the validity of a new paradigm:  informed navigation by LLM as an effective approach to reasoning based on knowledge graphs. This paradigm can be easily applied to various types of knowledge graphs.
%We analyze key design choices---function output formatting, few-shot exemplar quantity, and model selection---showing how careful design of these components improves overall performance and efficiency, providing practical guidance for LLM-based KG navigation systems.
\item Our study validates a new paradigm: LLM-driven navigation as an effective approach to knowledge graph reasoning. This paradigm also generalizes across different KG schemas.
\end{itemize}

\section{Related Work}
\label{sec:related-work}
We review several groups of approaches to KGQA.

\noindent \textbf{Semantic Parsing Methods. } 
Semantic parsing techniques transform natural language questions into executable logical forms before KG querying. RNG-KBQA \citep{ye2021rng} enumerates candidate logical forms through KG path searches, then employs ranking and generation models for executable form composition. A different approach is taken by DecAF \citep{yu2022decaf}, which linearizes KBs into text documents, enabling retrieval-based joint decoding of both logical forms and direct answers. FC-KBQA \citep{zhang2023fc} uses fine-to-coarse composition to address generalization and executability, reformulating KB components into middle-grained knowledge pairs.

\begin{sloppypar}
LLM-based methods have since emerged to leverage language models' capabilities for logical form generation. ChatKBQA \citep{luo2024chatkbqa} utilizes generate-then-retrieve pipelines, where instruction-tuned LLMs produce candidate logical forms subsequently grounded through phrase-level retrieval. In contrast, CoG \citep{zhao2025correcting} generates fact-aware queries through parametric knowledge output, then corrects hallucinated entities via KG alignment. DARA \citep{fang2024dara} introduces dual mechanisms for high-level task decomposition and low-level task grounding. Meanwhile, Rule-KBQA \citep{zhang2025rule} employs learned rules guiding generation through induction and deduction phases with symbolic agents. HTML \citep{wulamu2025html} proposes hierarchical multi-task learning with auxiliary tasks for entities, relations, and logical forms.
\end{sloppypar}

While these approaches provide interpretable traces and error recovery, they require generating complete logical forms or query plans upfront, demanding extensive schema knowledge and struggling when presumed schema elements are absent. In contrast, navigation based solely on locally available relations adapts better to actual KG structures without upfront schema requirements.

\noindent \textbf{Subgraph Retrieval Methods.}
These approaches involve first retrieving relevant graph portions around topic entities, then proceeds with reasoning over the induced subgraph. GRAFT-Net \citep{sun2018open} exemplifies early neural approaches by constructing heterogeneous subgraphs that merge KB entities with Wikipedia text, utilizing graph networks with directed propagation for multi-hop inference. PullNet \citep{sun2019pullnet} employs iterative subgraph expansion using graph CNNs to determine which nodes to ``pull'' next. TransferNet \citep{shi2021transfernet} transfers entity scores along activated edges through attention mechanisms while attending to question spans.

More sophisticated retrieval strategies have been proposed to address coverage and noise issues. UniKGQA \citep{jiang2022unikgqa} uses question-relation score propagation along KG edges for unified retrieval-reasoning. SR+NSM \citep{zhang2022subgraph} employs trainable subgraph retrievers decoupled from reasoning to enable plug-and-play enhancement. CBR-SUBG \citep{das2022knowledge} dynamically retrieves similar k-NN training queries with structural similarity. G-Retriever \citep{he2024g} formulates subgraph selection as Prize-Collecting Steiner Tree problems, while EPR \citep{ding2024enhancing} models structural dependencies through atomic adjacency patterns. Paths-over-Graph \citep{tan2025paths} uses multi-hop path expansion with graph reduction and pruning.

% These methods face fundamental trade-offs: larger subgraphs boost recall but introduce noise, while smaller ones risk missing critical edges. Furthermore, subgraph extraction and pruning are performed using criteria that are different from the reasoning logic. Typically, a semantic similarity with the question is used for the selection of entities and relations. An entity that appears dissimilar to the question, but is useful for the reasoning, may be filtered out. Reasoning may fail due to the wrong selection of entities in the subgraph. 
%answer quality is solely dependent on retrieval completeness—key relations filtered during construction cannot be recovered by reasoning modules.
These methods face a fundamental trade-off: larger subgraphs improve recall but introduce noise, while smaller ones risk missing critical edges. More importantly, subgraph extraction and pruning rely on criteria disconnected from the actual reasoning requirements. Typically, entities and relations are selected based on semantic similarity to the question, which may filter out logically necessary but lexically dissimilar edges. When such edges are pruned during subgraph construction, reasoning fails regardless of the quality of the downstream reasoning module.

\noindent \textbf{Agentic LLM Methods. }
This paradigm is characterized by interactive KG exploration through LLM agents. Think-on-Graph \citep{sun2024thinkongraph} performs iterative beam search maintaining top-$N$ partial paths, pruning candidates by scoring relations based on semantic similarity to the original question--without considering the reasoning history that led to the current entity. Plan-on-Graph \citep{chen2024plan} decomposes questions into sub-objectives and uses memory and reflection mechanisms to adapt exploration breadth and correct erroneous paths. ReKnoS \citep{wang2025reasoning} introduces super-relations enabling bidirectional reasoning and reducing misdirection errors. ORT \citep{liu2025ontology} reverses the typical exploration direction, using KG ontology to construct reasoning paths from purpose labels back to condition labels. Debate-on-Graph \citep{ma2025debate} employs multi-role LLM debate to iteratively simplify complex questions into single-hop subproblems. Spinach \citep{liu2024spinach} iteratively constructs SPARQL queries, learning KG structure from intermediate execution results. Readi \citep{cheng2024call} generates reasoning paths upfront and invokes LLM editing when KG instantiation fails. Generate-on-Graph \citep{xu2024generate} handles incomplete KGs by generating missing triples from LLM parametric knowledge when traversal fails.

While these single-model approaches demonstrate the potential of LLM-driven exploration, multi-model architectures have emerged to balance planning and efficiency. EffiQA \citep{dong-etal-2025-effiqa} uses an LLM for global planning and self-reflection while offloading semantic pruning to a small plug-in model for efficient KG exploration. KELDaR \citep{li2024framework} decomposes questions into atomic sub-questions and uses retriever-reranker models for targeted two-hop KG retrieval at each step. FiDeLiS \citep{sui2024fidelis} uses embedding models to pre-select candidate entities/relations via Path-RAG, then employs LLM-driven beam search with deductive verification to construct and validate reasoning paths step-by-step. iQUEST \citep{wang2025iquest} integrates iterative question decomposition with a GNN that aggregates 2-hop neighbor information for forward-looking entity selection. 

% All of these agentic methods do not fully exploit the capabilities of LLMs to their full potential. 
While these methods do not construct complete upfront queries, they often introduce complex multi-component architectures requiring separate modules for planning, memory, and pruning. Moreover, ToG maintains multiple parallel candidate paths in memory and uses beam search with similarity-based scoring to select among them. This adds complexity and memory overhead while risking incorrect path selection, since similarity to the question does not guarantee reasoning validity. %exponentially expands the search space, inundating LLMs with extraneous information.

Our assumption is that all of these tasks can be very well performed by a single LLM. Thus, our approach also falls into the group of agentic methods, but differs from other agentic approaches on the following aspects: (1) SoG does not decompose the question independently from the KG; (2) SoG does not maintain a subgraph in memory, and instead interacts iteratively with the KG; (3) The LLM performs path selection directly, informed by the full reasoning history and current KG neighborhood.

\section{Preliminaries}

\subsection{Knowledge Graphs}

A knowledge graph (KG) $\mathcal{G} = \{(e, r, e') \mid e, e' \in \mathcal{E}, r \in \mathcal{R}\}$ represents structured factual knowledge, where $\mathcal{E}$ and $\mathcal{R}$ denote the entity and relation sets, respectively. Each triple $(e, r, e')$ encodes a factual relationship $r$ between head entity $e$ and tail entity $e'$. Entities are uniquely identified by specific IDs (e.g., \texttt{m.07\_m2} represents Vincent van Gogh in Freebase) and may possess associated textual labels and semantic types for human interpretation.
For any entity $e$, its neighborhood structure comprises both outgoing and incoming relations. We formally define the neighboring relations as $\mathcal{R}_e = \{r \mid (e, r, e') \in \mathcal{G}\} \cup \{r \mid (e', r, e) \in \mathcal{G}\}$, encompassing relations where $e$ serves as either subject or object. This bidirectional connectivity enables flexible traversal during reasoning, allowing navigation in either direction along relational edges.

\subsection{Reasoning Path}

Multi-hop reasoning over KGs requires constructing connected sequences of triples that link topic entities to answer entities. A reasoning path $\mathcal{P}$ of length $k$ from entity $e_0$ to entity $e_k$ is formally defined as:
$$\mathcal{P} = [(e_0, r_1, e_1), (e_1, r_2, e_2), \ldots, (e_{k-1}, r_k, e_k)]$$ 
where each consecutive pair of triples shares an entity, creating a connected traversal through the graph structure. Intermediate entities $e_1, \ldots, e_{k-1}$ serve as stepping stones.

For the example illustrated in Figure~\ref{fig:sog-workflow}, the reasoning path is: Vincent van Gogh $\xrightarrow{\text{Place of birth}}$ Zundert $\xrightarrow{\text{Contained by}}$ Netherlands $\xrightarrow{\text{Capital}}$ Amsterdam. This 3-hop path demonstrates how complex questions requiring decompositional reasoning can be decomposed into sequential relational steps, each building upon the previous entity to reach the final answer.

\subsection{Knowledge Graph Question Answering}

Knowledge Graph Question Answering (KGQA) addresses the challenge of answering natural language questions using structured knowledge stored in a KG. Given a natural language question $q$, a KG $\mathcal{G}$, and topic entities $\mathcal{T}_q \subseteq \mathcal{E}$ mentioned in $q$, the objective is to identify answer entities $\mathcal{A}_q \subseteq \mathcal{E}$. A first required task is entity linking, which connects mentions in natural language questions with entity IDs in KG. As per prior work \citep{luo2024reasoning, sun2024thinkongraph, chen2024plan}, we use the gold entity linking provided in the datasets, and focus on the subsequent reasoning process.
%where entity mentions in questions are already linked to their KG identifiers. %, thus bypassing the need for entity linking.

\section{Methodology}
\label{sec:methodology}
SoG interacts with the KG iteratively to obtain the connected entities and relations from a given entity. Based on this, the LLM will choose the most relevant ones. We start by describing the search function.

\subsection{The \textsc{Search} Function}
\label{sec:search_function}

The \textsc{Search} function (see Algorithm~\ref{alg:search}) returns the 1-hop neighbours of a given entity in a specified direction. This function serves as the LLM's sole interface for KG exploration via tool calls, accepting three parameters:

\begin{itemize}
    \item \texttt{entity}: The given entity identifier (e.g., \texttt{m.07\_m2} for Vincent van Gogh)
    \item \texttt{direction}: Either \texttt{outgoing} (entity as subject) or \texttt{incoming} (entity as object)  
    \item \texttt{properties} (optional): Specific properties to filter results for focused exploration
\end{itemize}

The function returns results in a space-efficient markdown table format, prefixed with a row count that provides the LLM with immediate context about the result size. Each row contains four columns---property ID, property label, value ID, and value label---providing both machine-readable identifiers and human-readable labels. As demonstrated in Figure~\ref{fig:sog-workflow}, calling the function to get Vincent van Gogh's outgoing neighbours returns:

% \begin{figure*}[t]
\begin{functionoutput1}
594 rows:\\[1mm]
\begin{tabular}{l|l|l|l}
property & propertyLabel & value & valueLabel \\
\hline
people.person.profession & Profession & m.0n1h & Artist \\
visual\_art.visual\_artist.art\_forms & Art forms & m.05qdh & Painting \\
people.person.place\_of\_birth & Place of birth & m.0vlxv & Zundert \\
people.person.date\_of\_birth & Date of birth & 1853-03-30 & - \\
% visual\_art.visual\_artist.artworks & Artworks & m.0479\_q & The Starry Night \\
... & ... & ... & ...
\end{tabular}
\end{functionoutput1}
% \end{figure*}

\subsection{Handling High-Degree Nodes}
\label{sec:high_degree_handling}

KGs often contain high-degree nodes---entities with up to thousands %or even millions 
of connections such as countries, celebrities, or major organizations. Naively retrieving all neighbours of such nodes would overwhelm the LLM's context window and introduce excessive noise. We address this through a two-stage filtering mechanism formalized in Algorithm~\ref{alg:search}.

\begin{algorithm}%[h]
\caption{\textsc{Adaptive neighbourhood retrieval}}
\label{alg:search}
\KwIn{\texttt{entity}, \texttt{direction}, \texttt{properties}; thresholds $k, p$}
\KwOut{1-hop neighbours of \texttt{entity} in markdown table format}
\BlankLine
% $R \gets \Get(\texttt{entity}, \texttt{direction}, \texttt{properties})$\;
$R \gets \Get(\texttt{entity}, \texttt{direction},$\\
\hspace*{11.7em}$\texttt{properties})$\;
\BlankLine
\If{$|R| > k$ \textbf{and} \textnormal{\texttt{properties}} is empty}{
  $U \gets \Extract(R)$\;
  \Return \Format(U)\;
}
\If{$|R| > p$}{
  $R \gets R[0{:}p]$\;
}
\Return \Format($R$)\;
\end{algorithm}
% \vspace{-1em}

When the function encounters an entity with more than $k$ connected neighbours $R$ without specified properties, our function returns only the unique properties $U$ rather than all neighbour instances. As shown in Figure~\ref{fig:sog-workflow}, querying for the \textit{Netherlands} outgoing neighbours returns:
% \vspace{-2.5em}
\begin{functionoutput2}
% \footnotesize
% 60 rows:\\[2mm]
\begin{tabular}{l|l}
property & propertyLabel \\
\hline
location.country.form\_of\_government & Form of government \\
location.country.official\_language & Official language \\
location.country.capital & Capital \\
... & ...
\end{tabular}
\end{functionoutput2}

% \vspace{1em}
This property-only view allows the LLM to first survey available relation types without context overflow. The LLM then makes a targeted second call using the \texttt{properties} parameter to retrieve only relevant relations. This transforms high-degree node navigation from an intractable problem into two manageable steps: property discovery followed by selective retrieval. 

% Even with filtering, results may exceed practical limits. So, it is necessary to perform some selection in practice.
% Algorithm~\ref{alg:search} includes a selection step that selects the first $p$ triples ($|R|>p$) to ensure the response fits within context limits. 
% $p$ is set at 1000 in our experiments, which is large enough to include the relevant entities and relations in most cases.
Even with property filtering, results may exceed practical limits. Algorithm~\ref{alg:search} addresses this by truncating to the first $p$ triples when $|R|>p$, ensuring the response fits within context limits. We set $p=1000$ in our experiments, which is large enough to include the relevant entities and relations in most cases.

% \subsection{Search-on-Graph Prompting}
% \label{sec:sog_prompting}

% We guide the LLM's navigation strategy through few-shot prompting, with navigation exemplars that demonstrate three key aspects of effective KG traversal:
% \begin{itemize}
%     \item \textbf{Initial exploration}: making the first \textsc{Search} call based on the question's focus.
%     \item \textbf{Iterative traversal}: analyzing retrieved neighbours, selecting the relevant relations, and chaining \textsc{Search} calls to construct reasoning paths.
%     \item \textbf{Answer extraction}: recognizing when sufficient information has been gathered and extracting the final answer.
% \end{itemize}

% These exemplars demonstrate to the LLM how to navigate the KG through systematic observation and decision-making. The resulting traces remain fully interpretable as each navigation step is explicitly recorded through tool calls. Appendix~\ref{app:tool_defs} provides tool definitions, Appendix~\ref{app:tool_instructions} provides the full instructions given to the LLM, and Appendix~\ref{app:exemplar_samples} provides one representative exemplar per dataset.

\subsection{Search-on-Graph Prompting}
\label{sec:sog_prompting}
% We guide the LLM's navigation strategy through few-shot prompting with exemplars that demonstrate systematic KG traversal. Each exemplar illustrates three phases:

We guide the LLM's navigation through a system prompt that defines the task and tool usage, combined with five manually constructed few-shot exemplars demonstrating effective traversal patterns.

\noindent \textbf{System Instructions.}
The system prompt establishes the LLM's role and core navigation behavior:

\begin{promptbox}
You are a question-answering agent specializing in knowledge-graph question answering. You will receive a question and may call a tool to navigate the knowledge graph, collect information, and then formulate an answer.

You may call search(entity, direction) to retrieve adjacent relations and 1-hop neighbouring entities. Continue making tool calls until you arrive at a final answer. Then, and only then, stop making tool calls and provide your answer.
\end{promptbox}

\noindent This instruction is central to our approach: the LLM autonomously judges when sufficient information has been gathered, rather than relying on external termination criteria or fixed path lengths.

\noindent \textbf{Few-shot Exemplars.}
The exemplars demonstrate three phases of KG traversal:

\textit{Initial exploration}---identifying the starting entity and making the first search call:
\begin{promptbox}
<entity\_id> is the topic entity. Look for outgoing relations from <entity\_id> related to <question\_focus>.\\
search(entity="\texttt{<entity\_id>}", direction="outgoing")
\end{promptbox}

% \paragraph{Initial Exploration.} Given a question and its linked topic entities, the LLM identifies which entity to explore first and issues a \textsc{Search} call. This is  illustrated in the following example:

% \begin{promptbox}
% % Question: Where did the continental celtic languages originate? \{`Continental Celtic languages': `m.06v3q8'\}\\
% % m.06v3q8 is the topic entity of the question. Look for outgoing relations from m.06v3q8.\\
% % Tool Call: [\{"id": "some\_unique\_id", "type": "function", "function": \{"name": "search","arguments": \{"entity": "m.06v3q8", "direction": "outgoing"\}\}\}]
% Question: <question>\\
% <entity\_id> is the topic entity of the question. Look for outgoing relations from <entity\_id>.\\
% Tool Call: [\{"id": "some\_unique\_id", "type": "function", "function": \{"name": "search", "arguments": \{"entity": "<entity\_id>", "direction": "outgoing"\}\}\}]
% \end{promptbox}

\textit{Iterative traversal}---analyzing returned neighbours and deciding the next action:
\begin{promptbox}
We see \texttt{<propertyLabel>} points to \texttt{<valueLabel>}. To verify \texttt{<constraint>}, check outgoing relations from \texttt{<value>}...\\
search(entity="\texttt{<value>}", direction="outgoing")
\end{promptbox}

% \paragraph{Iterative Traversal.} After each search returns neighbouring entities and relations, the LLM analyzes the results and decides whether further exploration is needed. The exemplar shows how to identify relevant relations from the returned neighbors:

% \begin{promptbox}
% % Suppose it returns:\\
% % property|propertyLabel|value|valueLabel\\
% % --|--|--|--\\
% % language.language\_family.member\_of\_language\_families|member of language families|m.01sd8|Celtic languages\\
% % language.language\_family.geographic\_distribution|geographic distribution|m.02j9z|Europe\\
% % kg.object\_profile.prominent\_type||language.language\_family|Language Family\\
% % We see there is a language.language\_family.geographic\_distribution (geographic distribution) property that points to the object m.02j9z (Europe).
% Suppose it returns:\\
% property|propertyLabel|value|valueLabel\\
% --|--|--|--\\
% property1|propertyLabel1|value1|valueLabel1\\
% property2|propertyLabel2|value2|valueLabel2\\
% property3|propertyLabel3|value3|valueLabel3\\
% We see propertyLabel1 points to valueLabel1, which ...
% \end{promptbox}

\textit{Answer extraction}---recognizing when sufficient evidence exists:
\begin{promptbox}
We see \texttt{<property>} confirms \texttt{<condition>}. This answers the question.
Final answer: \{\texttt{<answer>}\}
\end{promptbox}

% For multi-hop questions, this phase involves chaining multiple \textsc{Search} calls, with the LLM selecting which entities to expand based on the question requirements and the accumulated reasoning history.

% \paragraph{Answer Extraction.} The LLM recognizes when sufficient evidence has been gathered and synthesizes the final answer:

% \begin{promptbox}
% Final answer:\{<answer>\}.
% \end{promptbox}

Crucially, the LLM makes all navigation decisions based on the complete reasoning context (the question, the search history, and the currently available neighbours), rather than relying on separate retrieval or scoring modules. The resulting traces remain fully interpretable as each step is explicitly recorded through tool calls. Appendix~\ref{app:tool_defs} provides the tool definition.
% , Appendix~\ref{app:tool_instructions} provides the full instructions given to the LLM, and Appendix~\ref{app:exemplar_samples} provides representative exemplars for each dataset.

\section{Experiments}
\label{sec:experiments}

\subsection{Experimental Setup}
\label{sec:exp-setup}

\noindent\textbf{Datasets and Evaluation Metric.} We evaluate SoG on six KGQA benchmarks spanning two major knowledge graphs, Freebase \citep{bollacker2008freebase} and WikiData \citep{vrandevcic2014wikidata}. For Simple Questions (SimpleQA) \citep{bordes2015large}, WebQuestionsSP (WebQSP) \citep{yih2016value}, ComplexWebQuestions (CWQ) \citep{talmor2018web}, and GrailQA \citep{gu2021beyond}, we use Freebase. For QALD-9 \citep{perevalov2022qald} and QALD-10 \citep{perevalov2022qald}, we use Wikidata. Table~\ref{tab:dataset_stats} provides detailed statistics. For SimpleQA and GrailQA, we evaluate on the same 1,000-sample test subset adopted by ToG \citep{sun2024thinkongraph} to manage computational costs while enabling direct comparison with prior work. For other datasets, we use the full test sets. As per prior work \citep{li2023few, sun2024thinkongraph,chen2024plan}, we report exact match accuracy (Hits@1).

% \begin{table}[h]
% \centering
% \caption{Detailed statistics of KGQA datasets. * indicates we use the 1,000-sample test subset from \citet{sun2024thinkongraph}.}
% \label{tab:dataset_stats}
% \resizebox{\columnwidth}{!}{
% \begin{tabular}{lccccccc}
% \toprule
% \textbf{Dataset} & \textbf{KG} & \textbf{Answer Type} & \textbf{Train} & \textbf{Test} & \textbf{License} \\
% \midrule
% Simple Questions* & Freebase & Entity & 14,894 & 1,000 & CC License \\
% WebQSP & Freebase & Entity/Number & 3,098 & 1,639 & CC License \\
% CWQ & Freebase & Entity & 27,734 & 3,531 & -- \\
% GrailQA* & Freebase & Entity/Number & 44,337 & 1,000 & -- \\
% QALD-9 & Wikidata & Entity/Number & 371 & 136 & MIT \\
% QALD-10 & Wikidata & Entity/Number & 412 & 394 & MIT \\
% \bottomrule
% \end{tabular}
% }
% \end{table}

\noindent\textbf{Backbone Model.} We evaluate SoG using GPT-4o, accessed through the OpenAI API. We chose GPT-4o over GPT-4 due to its substantially lower cost, enabling evaluation across all six benchmarks at scale. SoG is designed as a plug-and-play framework compatible with any LLM that supports tool calling, requiring no fine-tuning or task-specific adaptation.

% \noindent\textbf{SoG Prompting.} For each dataset, we manually construct five few-shot exemplars from training set questions, covering diverse reasoning patterns including single-hop retrieval, multi-hop traversal, constraint verification, and aggregation.

\noindent\textbf{Baselines and Parameters.} We compare SoG with 22 baselines, grouped into subgraph retrieval methods, LLM baselines, and agentic LLM methods. The results for most of the baselines are taken from their original papers when no open-source code is available for reproduction. Semantic parsing methods are excluded due to their reliance on task-specific fine-tuning, which is orthogonal to our training-free paradigm. For SoG, we set the high-degree threshold $k=50$ and the maximum result size $p=1000$. %, balancing information completeness with context window constraints.

\begin{table*}[t]
\centering
\caption{Exact match accuracy (\%) of KGQA methods across six benchmarks. Bold and underlined values indicate best and second-best results per dataset, respectively. Datasets are grouped by underlying KG: Freebase and Wikidata.}
\label{tab:results}
% \resizebox{\textwidth}{!}{%
\begin{tabular}{lcccccc}
\toprule
\multirow{2}{*}{\textbf{Method}} & \multicolumn{4}{c}{\textbf{Freebase}} & \multicolumn{2}{c}{\textbf{Wikidata}} \\
\cmidrule(lr){2-5} \cmidrule(lr){6-7}
 & \textbf{SimpleQA} & \textbf{WebQSP} & \textbf{CWQ} & \textbf{GrailQA} & \textbf{QALD-9} & \textbf{QALD-10} \\
 % & & & & & & \\
\midrule
\multicolumn{7}{c}{\textit{Subgraph Retrieval Methods}} \\
\midrule
GRAFT-Net \citep{sun2018open} & - & 66.4 & 32.8 & - & - & - \\
PullNet \citep{sun2019pullnet} & - & 68.1 & 47.2 & - & - & - \\
TransferNet \citep{shi2021transfernet} & - & 71.4 & 48.6 & - & - & - \\
UniKGQA \citep{jiang2022unikgqa} & - & 77.2 & 51.2 & - & - & - \\
% EWEK-QA + GPT-3.5 \citep{dehghan2024ewek} & 50.9 & 71.3 & 52.5 & 60.4 & - & - \\
SubgraphRAG + GPT-4o \citep{li2024simple} & - & 90.9 & 67.5 & - & - & - \\
\midrule
\multicolumn{7}{c}{\textit{LLM Baselines}} \\
\midrule
% IO Prompting + Qwen3-30B & 24.8 & 61.1 & 39.0 & 26.7 & 65.1 & 47.2 \\
% IO Prompting + Qwen3-235B & 30.3 & 61.1 & 51.0 & 32.3 & 62.7 & 48.7  \\
IO Prompt + GPT-4o & 34.3 & 68.4 & 51.7 & 35.3 & \underline{66.7} & 45.3 \\
Chain-of-Thought + GPT-4o & 32.6 & 69.0 & 52.4 & 38.0 & 65.1 & 44.8 \\
Self-Consistency + GPT-4o & 35.1 & 68.2 & 53.8 & 37.4 & 63.5 & 47.2 \\
\midrule
% \multicolumn{7}{c}{\textit{Semantic Parsing Methods}} \\
% \midrule
% DeCAF \citep{yu2022decaf} & - & 82.1 & 70.4 & - & - & - \\
% RE-KBQA \citep{cao2023pay} & - & 74.6 & 50.3 & - & - & - \\
% Interactive-KBQA \citep{xiong2024interactive} & - & 72.5 & 59.2 & - & - & - \\
% ChatKBQA \citep{luo2024chatkbqa} & - & 86.4 & 86.0 & - & - & - \\
% Reasoning-on-Graphs \citep{luo2024reasoning} & - & 85.7 & 62.6 & - & - & - \\
% Rule-KBQA \citep{zhang2025rule} & - & 84.1 & 73.5 & - & - & - \\
% RGR-KBQA \citep{feng2025rgr} & - & 84.5 & 82.0 & - & - & - \\
% KG-Agent \citep{jiang2025kg} & - & 83.3 & 72.2 & - & - & - \\
% HTML \citep{wulamu2025html} & - & 84.1 & 82.9 & - & - & - \\
% Correcting-on-Graphs \citep{zhao-etal-2025-correcting} & - & 83.3 & 63.9 & - & - & - \\
% CRF \citep{zhang2025collaborative} & - & 79.5 & 68.2 & - & - & - \\
% \midrule
\multicolumn{7}{c}{\textit{Agentic LLM Methods}} \\
\midrule
Think-on-Graph + GPT-4 \citep{sun2024thinkongraph} & 66.7 & 82.6 & 69.5 & 81.4 & - & 54.7 \\
Think-on-Graph + GPT-4o (reproduced) & 57.2 & 80.2 & 48.9 & 65.5 & 58.9 & 45.2 \\
Generate-on-Graph + GPT-4 \citep{xu2024generate} & - & 84.4 & \textbf{75.2} & - & - & - \\
Plan-on-Graph + GPT-4 \citep{chen2024plan} & - & 87.3 & 75.0 & \underline{84.7} & - & - \\
Readi + GPT-4 \citep{cheng2024call} & - & 78.7 & 67.0 & - & - & - \\
Spinach + GPT-4o \citep{liu2024spinach} & - & - & - & - & 58.3 & \underline{63.1} \\
FiDeLiS + GPT-4-Turbo \citep{sui2024fidelis} & - & 84.4 & 71.5 & - & - & - \\
FiDeLiS + GPT-4o \citep{sui2024fidelis} & - & 81.2 & 65.3 & - & - & - \\
% MindMap + GPT-4o \citep{wen-etal-2024-mindmap} & - & 61.2 & 51.3 & - & - & - \\
KELDaR + GPT-4 \citep{li2024framework} & - & 84.7 & 63.0 & - & - & - \\
EffiQA + GPT-4 \citep{dong-etal-2025-effiqa} & \underline{76.5} & 82.9 & 69.5 & 78.4 & - & 51.4 \\
ReKnoS + GPT-4 \citep{wang2025reasoning} & 69.3 & 84.9 & 68.2 & 82.7 & - & - \\
iQUEST + GPT-4o \citep{wang2025iquest} & - & 88.9 & 73.9 & 73.5 & - & - \\
ORT + GPT-4o \citep{liu2025ontology} & - & 87.7 & 65.4 & - & - & - \\
Debate-on-Graph + GPT-4 \citep{ma2025debate} & - & \underline{91.0} & 56.0 & 80.0 & - & - \\
% SRP + GPT-4.1-mini \citep{zhu2025self} & - & 83.6 & 69.0 & 78.8 & - & - \\
% KnowPath + DeepSeek-V3 \citep{zhao2025knowpath} & 65.3 & 89.0 & 73.5 & - & - & - \\
\midrule
% \multicolumn{7}{c}{\textit{SoG Prompting}} \\
% \midrule
% \textbf{SoG + Qwen3-30B (Ours)} & \underline{86.2} & 88.2 & 70.0 & 81.4 & \underline{81.0} & \underline{77.5} \\
% \textbf{SoG + Qwen3-235B (Ours)} & \textbf{86.4} & 89.3 & \textbf{77.1} & 83.9 & \textbf{82.5} & \textbf{79.8} \\
\textbf{SoG + GPT-4o (Ours)} & \textbf{84.8} & \textbf{91.3} & \underline{75.1} & \textbf{86.9} & \textbf{79.4} & \textbf{74.4} \\
% \midrule
% \multicolumn{7}{c}{\textit{SoG Supervised Fine-Tuned}} \\
% \midrule
% \textbf{SoG + Qwen3-30B-SFT (Ours)} & - & \textbf{91.3} & \textbf{78.8} & - & - & - \\
\bottomrule
\end{tabular}
% }
\end{table*}

\subsection{Main Results}
\label{sec:main_results}
% Table~\ref{tab:results} presents the performance of SoG and competing methods across all six benchmarks. Our approach consistently achieves state-of-the-art or highly competitive results without any task-specific fine-tuning. SoG + GPT-4o outperforms all prior systems on five of six datasets, trailing only Generate-on-Graph on CWQ by 0.1\%. The improvements over previous best methods range from incremental to substantial. On Freebase datasets, we improve by +8.3\% on SimpleQA (vs.\ EffiQA), +0.3\% on WebQSP (vs.\ Debate-on-Graph), and +2.2\% on GrailQA (vs.\ Plan-on-Graph). The improvements are particularly striking on Wikidata benchmarks, where we achieve double-digit gains: +12.7\% on QALD-9 (vs.\ IO Prompt + GPT-4o) and +11.3\% on QALD-10 (vs.\ Spinach).

Table~\ref{tab:results} presents the performance of SoG and competing methods across all six benchmarks. Our approach consistently achieves state-of-the-art or highly competitive results without any task-specific fine-tuning. SoG + GPT-4o outperforms all prior systems on five of six datasets, trailing only Generate-on-Graph on CWQ by 0.1\%. The improvements range from incremental on Freebase datasets (+0.3\% on WebQSP, +2.2\% on GrailQA, +8.3\% on SimpleQA) to substantial double-digit gains on Wikidata benchmarks (+12.7\% on QALD-9, +11.3\% on QALD-10).

% We analyze these results by examining how SoG differs from prior agentic methods along three dimensions: the use of global question decomposition, subgraph extraction with beam search, and access to full reasoning context.

The strong performance across both Freebase (SimpleQA, WebQSP, CWQ, GrailQA) and Wikidata (QALD-9, QALD-10) benchmarks validates our schema-agnostic design. While Freebase uses compound value types for complex relations and Wikidata employs qualifiers, SoG adapts to both structures without modification, confirming that our single function approach generalizes across different KG schemas. Furthermore, SoG is shown to be effective across both single-hop (SimpleQA) and multi-hop (WebQSP, CWQ, GrailQA, QALD-9, QALD-10) datasets. This contrasts with methods such as Think-on-Graph, EffiQA, and ReKnoS, which show stronger relative performance primarily on multi-hop tasks. Our consistent performance across complexity levels likely stems from giving the LLM access to the full reasoning history at each decision point. With this context, the LLM typically selects a single relation per hop, in contrast to beam search methods that must select multiple relations and maintain parallel exploration paths. This avoids the noise accumulation that can overwhelm simpler questions without sacrificing performance on complex reasoning chains.

% - Comparison with question decomposition approaches

% - Comparison with interactive methods with KG (ToG) - the difference: no subgraph and ask LLM to select, selection of relation/node with full reasoning history

Many baseline methods report results using GPT-4, while we evaluate with GPT-4o. To assess whether this comparison is fair, we reproduced ToG with GPT-4o under identical conditions. As shown in Table~\ref{tab:results}, ToG + GPT-4o substantially underperforms ToG + GPT-4 across all datasets. This aligns with broader observations that GPT-4 exhibits stronger reasoning capabilities than GPT-4o for complex tasks. Consequently, SoG + GPT-4o outperforms methods using the more capable GPT-4 backbone, showing that our gains stem from methodological improvements rather than model advantages.

\subsection{Effect of Backbone Models}
\label{sec:backbone}

To further investigate how SoG generalizes across different LLM backbones, we evaluate GPT-4o and three open-source Qwen3 variants: Qwen3-4B-Thinking-2507, Qwen3-30B-A3B-Thinking-2507, and Qwen3-235B-A22B-Thinking-2507-FP8 (abbreviated as Qwen3-4B, Qwen3-30B, and Qwen3-235B, respectively). Table~\ref{tab:backbone} presents results across all six benchmarks.

%To investigate how SoG generalizes across different LLM backbones, we evaluate GPT-4o and three open-source Qwen3 variants: Qwen3-4B-Thinking-2507, Qwen3-30B-A3B-Thinking-2507, and Qwen3-235B-A22B-Thinking-2507-FP8 (abbreviated as Qwen3-4B, Qwen3-30B, and Qwen3-235B, respectively). Table~\ref{tab:backbone} presents results across all six benchmarks.

\begin{table}[t]
\centering
\caption{Performance of SoG using different backbone models: open-source Qwen3 models and  proprietary model GPT-4o. Bold indicates best per dataset.}
\label{tab:backbone}
\resizebox{\columnwidth}{!}{%
\begin{tabular}{lcccccc}
\toprule
\textbf{Backbone} & \textbf{SimpleQA} & \textbf{WebQSP} & \textbf{CWQ} & \textbf{GrailQA} & \textbf{QALD-9} & \textbf{QALD-10} \\
\midrule
Qwen3-4B & 76.3 & 80.1 & 56.1 & 80.4 & 75.4 & 69.4 \\
Qwen3-30B & 86.2 & 88.2 & 70.0 & 81.4 & 81.0 & 77.5 \\
Qwen3-235B & \textbf{86.4} & 89.3 & \textbf{77.1} & 83.9 & \textbf{82.5} & \textbf{79.8} \\
GPT-4o & 84.8 & \textbf{91.3} & 75.1 & \textbf{86.9} & 79.4 & 74.4 \\
\bottomrule
\end{tabular}%
}
\end{table}

Larger models generally improve performance. Among the Qwen3 models, Qwen3-235B outperforms Qwen3-30B on all six datasets, with the most substantial gain on CWQ (+7.1\%), the most complex multi-hop benchmark. The jump from 4B to 30B is even more pronounced, with gains of +9.9\% on SimpleQA, +13.9\% on CWQ, and +8.1\% on both WebQSP and QALD-10. Qwen3-235B leads on four of six datasets (SimpleQA, CWQ, QALD-9, QALD-10), while GPT-4o leads on WebQSP and GrailQA. This demonstrates that SoG does not require proprietary models to achieve state-of-the-art performance.

Notably, even the smallest model achieves strong results compared to several baselines. Qwen3-4B surpasses all prior methods on both Wikidata benchmarks (75.4\% on QALD-9 vs.\ previous best 66.7\%; 69.4\% on QALD-10 vs.\ previous best 63.1\%) despite being a compact 4B dense model. On Freebase, Qwen3-4B also outperforms several GPT-4-based agentic methods. This demonstrates that SoG remains effective even with compact models, making it practical for cost-sensitive deployment scenarios. 

% Notably, even compact models achieve strong results with SoG. Qwen3-4B surpasses all prior methods on both Wikidata benchmarks (75.4\% on QALD-9 vs.\ previous best 66.7\%; 69.4\% on QALD-10 vs.\ previous best 63.1\%) despite being orders of magnitude smaller than GPT-4. On Freebase benchmarks, Qwen3-30B outperforms GPT-4-based methods including ToG (82.6\% on WebQSP), EffiQA (69.5\% on CWQ), and ReKnoS (68.2\% on CWQ). These results suggest that SoG's simple interface---a single \textsc{Search} function with context-aware navigation---effectively leverages LLMs' inherent reasoning capabilities without requiring the scale of frontier models. By presenting KG exploration as iterative observation and decision-making rather than complex multi-step planning, SoG aligns naturally with how LLMs reason, enabling even smaller models to outperform larger ones operating under more complex frameworks. This makes SoG practical for cost-sensitive and latency-critical deployment scenarios where proprietary APIs or massive models are infeasible.

\label{sec:fewshot-analysis}
\begin{figure*}[t]
    \centering
    \includegraphics[width=0.85\linewidth]{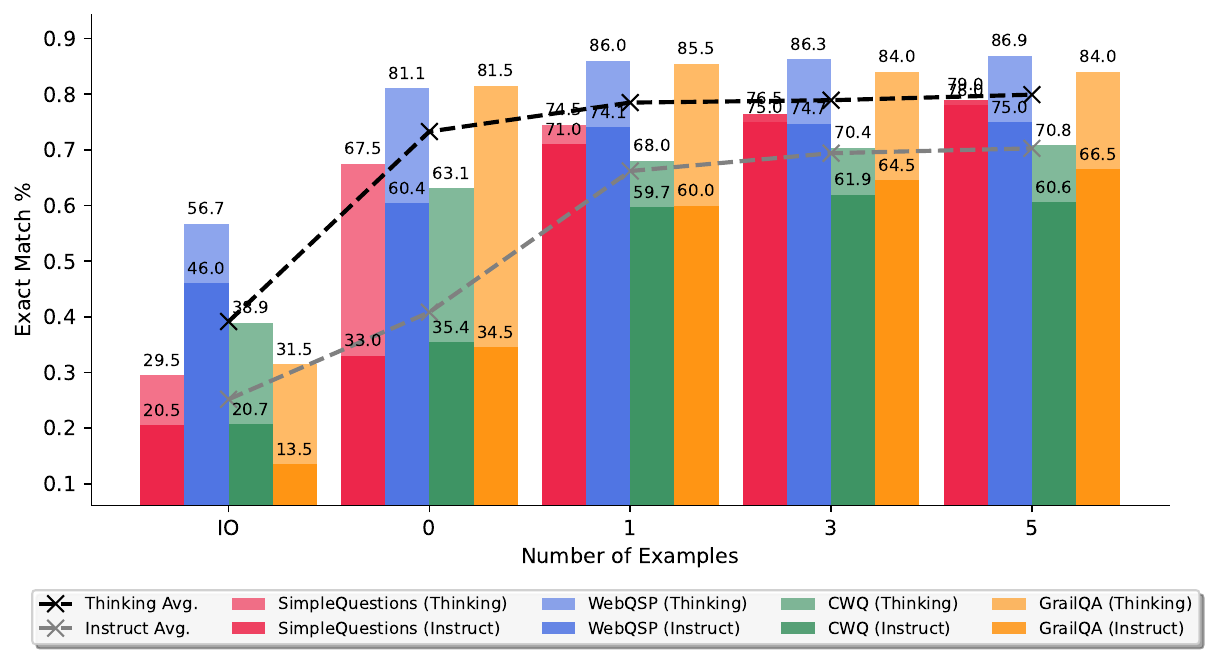}
    \small
    % \vspace{-1em}
    \caption{Impact of few-shot exemplar quantity on exact match accuracy (\%) across four Freebase datasets (SimpleQA, WebQSP, CWQ, GrailQA) for Qwen3-30B-A3B-Thinking-2507 and Qwen3-30B-A3B-Instruct-2507.}
    \label{fig:fewshot-ablation}
    \Description{Line charts showing exact match accuracy versus number of few-shot exemplars (0 to 5) for two models across four datasets.}

\end{figure*}

\subsection{Efficiency Analysis}
\label{sec:efficiency}

We also evaluate the computational efficiency of SoG compared to ToG, the most widely adopted agentic LLM baseline. Table~\ref{tab:efficiency} presents the average number of LLM calls, token usage, and latency per question using GPT-4o as the backbone for both methods.

\begin{table}[t]
\centering
\caption{Efficiency comparison between SoG and ToG using GPT-4o. All metrics are averaged per question.}
\label{tab:efficiency}
\resizebox{\columnwidth}{!}{%
\begin{tabular}{ccccccc}
\toprule
\textbf{Dataset} & \textbf{Method} & \textbf{LLM Calls} & \textbf{Input Tokens} & \textbf{Output Tokens} & \textbf{Time (s)} \\
\midrule
\multirow{2}{*}{WebQSP} & ToG & 8.4 & \textbf{8,569.3} & 902.1 & 31.4 \\
& SoG & \textbf{3.9} & 22,252.1 & \textbf{238.2} & \textbf{16.1} \\
\midrule
\multirow{2}{*}{CWQ} & ToG & 11.4 & \textbf{11,204.3} & 1,322.5 & \textbf{55.5} \\
& SoG & \textbf{5.2} & 54,589.3 & \textbf{398.4} & 77.0 \\
\midrule
\multirow{2}{*}{GrailQA} & ToG & 5.6 & \textbf{4,194.1} & 675.2 & 20.0 \\
& SoG & \textbf{3.0} & 14,951.7 & \textbf{139.9} & \textbf{6.5} \\
\bottomrule
\end{tabular}%
}
\end{table}

SoG uses at least 46.4\% fewer LLM calls and 69.9\% fewer output tokens across all datasets. This is due to SoG typically selecting a single relation per hop based on the full context, then making one navigation decision before retrieving all relevant entities along that relation. In contrast, ToG's higher output token consumption stems from its beam search, which requires the LLM to generate extensive scoring and pruning reasoning across parallel paths at each depth---first scoring candidate relations, then scoring candidate entities, with both steps repeated for every path at every hop. 

However, SoG consumes at least 2.6$\times$ more input tokens because it maintains the complete search history for context-aware navigation. SoG is faster on WebQSP (2.0$\times$) and GrailQA (3.1$\times$), but slower on CWQ (1.4$\times$). For WebQSP and GrailQA, the output token savings outweigh increased input processing. For CWQ, as it requires longer context windows due to its longer reasoning chains, the output token savings do not outweigh the increased input tokens.

\subsection{Performance by Question Type}
\label{sec:question_type}

\begin{table}[t]
\centering
\caption{Performance breakdown by question type on CWQ.}
\label{tab:question_type}
\resizebox{\columnwidth}{!}{%
\begin{tabular}{cccccc}
\toprule
\textbf{Method} & \textbf{All} & \textbf{Composition} & \textbf{Conjunction} & \textbf{Superlative} & \textbf{Comparative} \\
\midrule
ToG & 48.9 & 49.9 & 50.1 & 37.1 & 42.7 \\
\midrule
\textbf{SoG} & \textbf{75.1} & \textbf{77.6} & \textbf{74.8} & \textbf{56.9} & \textbf{75.6} \\
% \midrule
% \textit{Improvement} & +26.2 & +27.7 & +24.7 & +19.8 & +32.9 \\
\bottomrule
\end{tabular}
}
\end{table}

\begin{table*}[t]
\centering
\caption{Comparison of different output formats on SimpleQA (20\% sample) using Qwen3-30B-A3B-Thinking-2507 with 5 exemplars. We report the average number of main interaction tokens, average number of reasoning tokens, average number of total tokens, average number of turns, and exact match (EM) accuracy. ``Markdown + Property Filter'' denotes our concise format with an additional filtering round, which achieves the best accuracy and efficiency.}
% \vspace{1em}
% \resizebox{\columnwidth}{!}{%
\begin{tabular}{lccccc}
\toprule
\textbf{Format} & \textbf{Main Tokens} & \textbf{Reasoning Tokens} & \textbf{Total Tokens} & \textbf{Turns} & \textbf{EM} \\
\midrule
JSON            & 9312.2             & 2735.2               & 12047.3             & 3.06             & 76.5       \\
Markdown        & 6028.1             & 1953.7               & 7981.8              & 3.05             & 74.5       \\
Markdown + Property Filter (Ours)  & 3715.9             & 1906.6               & 5622.5              & 3.93             & 78.0       \\
\bottomrule
\end{tabular}
% }
\label{tab:output-format-comparison}
\end{table*}

To understand where SoG's improvements are most pronounced, we analyze performance by question type on CWQ. Table~\ref{tab:question_type} compares SoG against ToG using the same GPT-4o backbone.

SoG improves across all question types, with the largest gains on Comparative (+32.9\%) and Composition (+27.7\%) questions. These types require maintaining coherent multi-hop reasoning chains, where SoG's context-aware navigation prevents the reasoning drift that occurs when ToG's beam search diverges across parallel paths. Both methods show their lowest performance on Superlative questions, which require retrieving and comparing complete candidate lists from high-degree nodes.

\subsection{Ablation Studies and Analysis}
\label{sec:ablation}
We conduct a series of ablation studies to analyze key design choices in SoG, examining three factors: the impact of few-shot exemplar quantity, reasoning-optimized models versus standard instruction models, and different output formatting on performance. All ablations use 20\% of samples from each Freebase-based test set. We evaluate on both Qwen3-30B-A3B-Thinking-2507 and Qwen3-30B-A3B-Instruct-2507.

\paragraph{Effect of Few-shot Exemplars.}
Figure~\ref{fig:fewshot-ablation} shows the performance of Thinking and Instruct models across varying exemplar quantities. The black and grey dashed lines represent the average exact match accuracy across the four datasets for the Thinking and Instruct models, respectively. Both models show dramatic improvements when transitioning from IO prompting to 0-shot with tool definitions, demonstrating that LLMs can perform structured navigation once they understand the \textsc{Search} function interface. Adding a single navigation exemplar (1-shot) produces another substantial boost across all datasets---from simple single hop to complex multi-hop tasks---confirming that even a single demonstration benefits all complexity levels. Performance plateaus at 3-shot with minimal gains thereafter, indicating that a small set of diverse exemplars sufficiently demonstrates effective navigation strategies.

\begin{sloppypar}
\paragraph{Thinking vs. Non-Thinking Models.}
The Thinking variant consistently outperforms the Instruct variant across all settings in Figure~\ref{fig:fewshot-ablation}, with the gap most pronounced on multi-hop datasets (WebQSP, CWQ, GrailQA) compared to single-hop SimpleQA. This performance difference reveals that model architecture and inherent reasoning capabilities are critical for SoG's effectiveness. The reasoning-optimized model better leverages our iterative observation-decision framework---analyzing available relations and making informed navigation choices based on reasoning rather than question semantics or pattern-matching against exemplars. While both models benefit from additional exemplars, the Thinking variant extracts more value from navigation demonstrations, indicating that SoG's performance ceiling depends on the model's underlying capacity for structured reasoning over KGs.  
\end{sloppypar}

\paragraph{Output Format and Filtering.}
Table~\ref{tab:output-format-comparison} compares the impact of different output formats on performance and efficiency. While the original JSON format that the SPARQL execution returns yields strong accuracy, it uses significantly more tokens than the other two formats. Switching to Markdown format reduces token usage considerably, but slightly impacts accuracy. Our property filtering approach introduces an additional stage: when encountering high-degree nodes, we first retrieve available properties, then make a targeted second call with relevant properties only. Despite requiring additional turns, this strategy achieves the lowest total token usage while simultaneously delivering the highest accuracy. The efficiency gain stems from avoiding redundant information in dense neighborhoods, while the accuracy improvement suggests that focused retrieval helps the LLM identify relevant paths more effectively. These results highlight how careful output design choices critically impacts both computational cost and performance in LLM-based KGQA systems.

\section{Conclusion}

We present Search-on-Graph (SoG), a simple yet effective framework for knowledge graph question answering where a single LLM navigates a KG through an iterative 1-hop \textsc{Search} function. Unlike prior methods that rely on multi-component architectures, upfront planning, or context-free beam search, SoG performs context-aware navigation by maintaining the full reasoning history at each decision point. This \textit{observe-think-navigate} approach enables the LLM to make informed relation selections grounded in the logical chain of reasoning rather than semantic similarity heuristics. Experiments across six benchmarks spanning Freebase and Wikidata demonstrate that SoG achieves state-of-the-art results on five of six datasets, with particularly substantial gains on Wikidata benchmarks. This validates our assumption that LLMs can better perform the task of selecting relevant relations than specifically designed methods in the literature. 
Our analysis shows that SoG remains effective across model sizes---even a compact 4B model surpasses prior methods on Wikidata. SoG also reduces LLM calls by at least 46.4\% compared to Think-on-Graph. These results suggest that architectural simplicity, combined with careful context management and output formatting, can outperform more elaborate approaches. The generalizability of SoG, requiring no task-specific training and adapting seamlessly across KG schemas, validates observation-driven navigation as a promising direction for LLM-based structured reasoning.

% \clearpage

\bibliographystyle{ACM-Reference-Format}
\bibliography{references}
\balance
% \bibliography

\clearpage
\appendix

\onecolumn
\raggedbottom
\section{Datasets}
\label{app:datasets}

Table~\ref{tab:dataset_stats} provides detailed statistics for all evaluation datasets, extending the overview in Section~\ref{sec:exp-setup}.

\begin{table}[h]
\centering
\caption{Detailed statistics of KGQA datasets. * indicates we use the 1,000-sample test subset from \citet{sun2024thinkongraph}.}
\label{tab:dataset_stats}
% \resizebox{\columnwidth}{!}{
\begin{tabular}{lccccccc}
\toprule
\textbf{Dataset} & \textbf{KG} & \textbf{Answer Type} & \textbf{Train} & \textbf{Test} & \textbf{License} \\
\midrule
Simple Questions* & Freebase & Entity & 14,894 & 1,000 & CC License \\
WebQSP & Freebase & Entity/Number & 3,098 & 1,639 & CC License \\
CWQ & Freebase & Entity & 27,734 & 3,531 & -- \\
GrailQA* & Freebase & Entity/Number & 44,337 & 1,000 & -- \\
QALD-9 & Wikidata & Entity/Number & 371 & 136 & MIT \\
QALD-10 & Wikidata & Entity/Number & 412 & 394 & MIT \\
\bottomrule
\end{tabular}
% }
\end{table}

%=============================================
% APPENDIX B: SEARCH TOOL DEFINITIONS
%=============================================
% \onecolumn
\raggedbottom
\section{Search Tool Definition}
\label{app:tool_defs}

% We provide the complete tool definitions used for Freebase and Wikidata knowledge graphs.
We provide the tool definition used for the Freebase knowledge graph.

\subsection{Freebase Search Tool Definition}
\label{app:freebase_tool_defs}

\begin{lstlisting}[style=pythonstyle]
TOOLS_FREEBASE = [
    {
        "type": "function",
        "function": {
            "name": "search",
            "description": (
                "Build and execute a SPARQL query on Freebase that retrieves adjacent properties, property labels, values, and value labels in the specified direction for a given entity."
            ),
            "parameters": {
                "type": "object",
                "properties": {
                    "entity": {
                        "type": "string",
                        "description": "The entity (e.g., 'm.04yd0fh') whose adjacent relations and entities we want to fetch."
                    },
                    "direction": {
                        "type": "string",
                        "enum": ["incoming", "outgoing"],
                        "description": "Direction of relationship to consider."
                    },
                    "properties_to_filter_for": {
                        "type": "array",
                        "items": {"type": "string"},
                        "description": "Optional list of specific properties to filter by (e.g., ['people.person.place_of_birth', 'people.person.nationality'])."
                    }
                },
                "required": ["question", "entity", "direction"],
                "additionalProperties": False
            },
        }
    },
]
\end{lstlisting}

\end{document}